\documentclass[10pt,twocolumn,letterpaper]{article}

\usepackage{cvpr}              
\makeatletter
\@namedef{ver@everyshi.sty}{}
\makeatother

\usepackage{graphicx}
\usepackage{amsmath}
\usepackage{amssymb}
\usepackage{booktabs}
\usepackage{xcolor}
\usepackage{colortbl}
\usepackage{highlight}
\usepackage{multirow}

\usepackage[pagebackref,breaklinks,colorlinks]{hyperref}

\usepackage[capitalize]{cleveref}
\crefname{section}{Sec.}{Secs.}
\Crefname{section}{Section}{Sections}
\Crefname{table}{Table}{Tables}
\crefname{table}{Tab.}{Tabs.}

\DeclareMathOperator*{\argmax}{argmax}

\def\cvprPaperID{4928} 
\def\confName{CVPR}
\def\confYear{2023}

\newcommand{\model}{AVFormer}

\begin{document}

\title{\model: Injecting Vision into Frozen Speech Models for Zero-Shot AV-ASR}


\author{Paul Hongsuck Seo\hspace{1cm}Arsha Nagrani\hspace{1cm}Cordelia Schmid\\
Google Research\\
{\tt\small \{phseo, anagrani, cordelias\}@google.com}
}
\maketitle

\begin{abstract}
Audiovisual automatic speech recognition (AV-ASR) aims to improve the robustness of a speech recognition system by incorporating visual information. Training fully supervised multimodal models for this task from scratch, however is limited by the need for large labelled audiovisual datasets (in each downstream domain of interest). We present \model, a simple method for augmenting audio-only models with visual information, at the same time performing lightweight domain adaptation.
We do this by (i)~injecting visual embeddings into a frozen ASR model using lightweight trainable adaptors. We show that these can be trained on a small amount of weakly labelled video data with minimum additional training time and parameters. (ii)~We also introduce a simple curriculum scheme during training which we show is crucial to enable the model to jointly process audio and visual information effectively; and finally (iii)~we show that our model achieves state of the art zero-shot results on three different AV-ASR benchmarks (How2, VisSpeech and Ego4D), while also crucially preserving decent performance on traditional audio-only speech recognition benchmarks (LibriSpeech). Qualitative results show that our model effectively leverages visual information for robust speech recognition.
\end{abstract}


\section{Introduction}
\label{sec:intro}
\begin{figure}
    \centering
    \includegraphics[width=\linewidth]{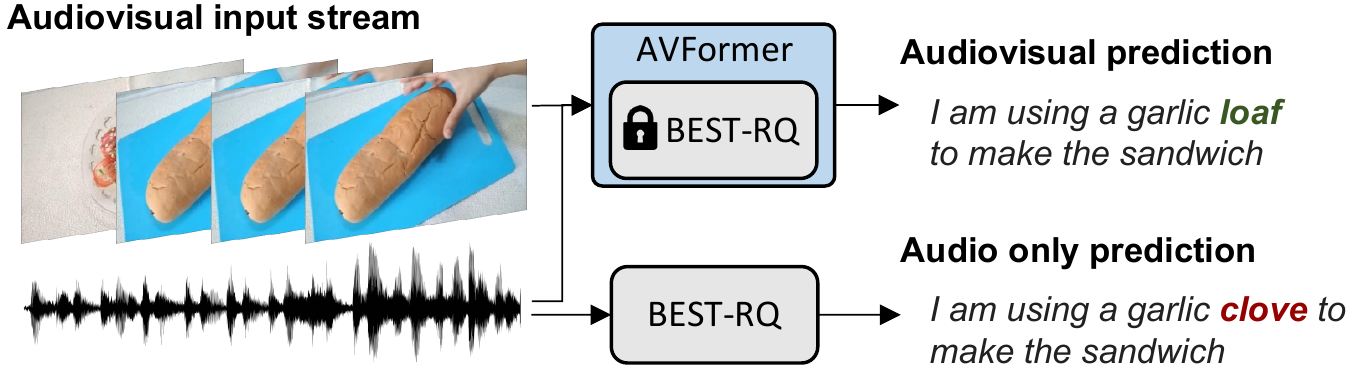}
    \caption{\textbf{Unconstrained audiovisual speech recognition.} We inject vision into a frozen speech model (BEST-RQ, in grey) for zero-shot audiovisual ASR via lightweight modules to create a parameter and data efficient model called AVFormer (blue).  
    The visual context can provide helpful clues for robust speech recognition especially when the audio signal is noisy (the visual loaf of bread helps correct the audio-only mistake \textcolor[RGB]{155,0,0}{clove} to \textcolor[RGB]{0,110,51}{loaf} in the generated transcript). } 
    \label{fig:av-asr}
\end{figure}

Robustness or adaptation to new, unconstrained domains is a key challenge for automatic speech recognition (ASR) systems. In multimodal video (\eg, TV, online edited videos), the visual stream can
provide strong cues for improving the robustness of ASR systems, particularly in cases
where the audio is noisy -- this is called audiovisual ASR (AV-ASR). Unlike works that simply focus on lip motion~\cite{Noda2014AudiovisualSR,Tamura2015AudiovisualSR,Chung2017LipRS,Afouras2018DeepAS,Petridis2018EndtoEndAS,Makino2019RecurrentNN,Ma2021EndToEndAS,Serdyuk2021AudioVisualSR}, we investigate the contribution of entire visual frames.
This is particularly useful for videos `in the wild', where the
mouth is not necessarily visible (\eg, egocentric viewpoints, face coverings, and low resolution etc.)~\cite{gabeur2022avatar}. 
The task is illustrated in Figure~\ref{fig:av-asr}.

Building audiovisual datasets for training AV-ASR models, however, is challenging. Datasets such as How2~\cite{sanabria2018how2} and VisSpeech~\cite{gabeur2022avatar} have been created from instructional videos online,
but they are small in size. 
Not only are datasets for this task small, but models are typically large and consist of both visual and audio encoders. For example the latest AV-ASR model AVATAR~\cite{gabeur2022avatar} shows impressive performance on both datasets, but requires the end-to-end training of visual and audio components in tandem, and consequently a large amount of compute. Like other AV-ASR works~\cite{Miao2016OpenDomainAS,Gupta2017VisualFeatures,Palaskar2018EndtoendMS,Caglayan2019VAT,Srinivasan2020LookingEL}, it is also only trained and tested on instructional videos, and as we show in the experiments, generalizes poorly to new domains in the zero-shot setting. 

On the other hand, there have been a number of recently released large-scale audio-only models~\cite{chiu2022self,chung2021w2v,hsu2021hubert} that are heavily optimised via self-supervised pretraining and large-scale supervised training on \textit{audio-only} data obtained from audio books such as LibriLight~\cite{kahn2020libri} and LibriSpeech~\cite{panayotov2015librispeech}. These models contain billions of parameters, are readily available, and show strong \textit{generalization across domains}.

Our goal is to reuse the extensive expertise and training time that has been invested in such models, by using their weights. We are inspired by recent works adapting \textit{frozen} foundation models for multi-modal tasks. A popular example is~\cite{alayrac2022flamingo} that injects visual information into large language models (LLMs) for vision-text tasks. The benefit of building on strong frozen LLMs for these tasks is the hope that this will enable the visual-text model to retain powerful \textit{language-only} abilities such as few-shot language adaptation or external knowledge retrieval. Our goal is simple - we wish to do the same for AV-ASR, using strong audio-only ASR models. We add visual inputs to these models in a lightweight manner to enable AV-ASR, but still maintain the benefits of audio-only pretraining for zero-shot generalization. 

Our framework is called \model{}, and injects visual information into a frozen ASR model using lightweight projection layers and trainable adaptors. We show that these can be trained on a small amount of weakly labelled video data (only 5\% of the data used by existing state of the art methods~\cite{gabeur2022avatar}) with minimum additional training time and parameters, minimizing the domain shift and catastrophic forgetting that can accompany end-to-end finetuning. In order to further ensure stability during the finetuning of these adapters, we also introduce a simple curriculum scheme during training which we show is crucial to enable the model to jointly process audio and visual information effectively. Finally,  we show that our model outperforms existing state of the art zero-shot methods on three different AV-ASR benchmarks (How2, VisSpeech and Ego4D) across different domains, while also crucially preserving decent performance on traditional audio-only speech recognition benchmarks (LibriSpeech). 


\section{Related Works}
\label{sec:related}
\noindent\textbf{State-of-the-Art Speech Recognition}
Recent state-of-the-art ASR models~\cite{chiu2022self,hsu2021hubert,chung2021w2v,zhang2020pushing,xu2021self} almost all adopt transformer based audio encoders~\cite{gulati2020conformer,synnaeve2019end,hsu2021hubert} embedding input audio signals into a set of token features thereby extracting local information within a temporal window.
Encoders are trained end-to-end using losses such as CTC~\cite{graves2006connectionist}, RNN-T~\cite{graves2012sequence} and LAS~\cite{chan2016listen}.
In many cases, these encoders are pretrained~\cite{chiu2022self,hsu2021hubert,chung2021w2v,zhang2020pushing,xu2021self} on large-scale unannotated datasets such as LibriLight~\cite{kahn2020libri}, and then finetuned for downstream ASR. Consequently, such models incorporate a number of highly-engineered training tricks and techniques suitable for ASR, which we want to reuse for multimodal inference.
Rebuilding a multimodal model from scratch incorporating these learnings is expensive and must be redone for each new model. As models get larger and larger~\cite{hsu2021hubert,chung2021w2v,zhang2020pushing}, this requires a prohibitive amount of compute. Our goal is to reuse this knowledge in a lightweight manner by injecting visual understanding capability into a readily available state-of-the-art ASR model. \\
\noindent\textbf{Audiovisual Speech Recognition }
Most AV-ASR works are focused on lip motion, right from early works that use pre-extracted features~\cite{Noda2014AudiovisualSR,Tamura2015AudiovisualSR} to more recent end-to-end approaches that work on pixels directly~\cite{Chung2017LipRS,Afouras2018DeepAS,Petridis2018EndtoEndAS,Makino2019RecurrentNN,Ma2021EndToEndAS,Serdyuk2021AudioVisualSR}. In contrast, the setting explored in this work is full frame AV-ASR beyond the speaker's mouth movements (also known as `context-aware' speech recognition). Here the defacto strategy is to use pre-extracted visual context features (due to the high dimensionality of full frame video) -- either action features~\cite{sanabria2018how2,Caglayan2019VAT,Ghorbani2021LLD,paraskevopoulos2020multires}, or place and object features~\cite{Miao2016OpenDomainAS,Gupta2017VisualFeatures,Palaskar2018EndtoendMS,Caglayan2019VAT,Srinivasan2020LookingEL}. Unlike these works which all use visual features from classification models trained on a closed-set of pre-defined objects, places or actions, we use features from CLIP~\cite{Radford2021CLIP}, which is trained on image and text paired data, and known to have strong generalization and zero-shot capabilities. This makes our features more suited to unconstrained videos `in the wild'. An outlier is the recently proposed AVATAR~\cite{gabeur2022avatar}, which uses full frame pixels and trains end-to-end on HowTo100. It is the state of the art for this task, achieving good performance on How2 and introducing a new dataset called VisSpeech. Unlike AVATAR, our method reuses strong frozen pretrained models, thereby requiring only 5\% of the audiovisual data used in AVATAR, and generalises much better across different domains in the zero-shot setting. \\ 
\noindent\textbf{Adapting Large Frozen Pretrained Models} There has been a recent flurry of works that adapt frozen foundation models for multi-modal tasks, most notably for injecting visual information to large language models (LLMs)~\cite{alayrac2022flamingo}.  Architectural details vary: for example MAGMA~\cite{eichenberg2021magma} and Frozen-BiLM~\cite{yang2022frozenbilm} add bottleneck adapters~\cite{houlsby2019parameter,sung2022vl} to the frozen LLM injecting some visual information; ClipCap~\cite{mokady2021clipcap} learns a vision-to-prefix bridging transformer to map vision features into a prefix for GPT-2, while VC-GPT~\cite{luo2022vc} adds new learnt layers to the frozen LLM. In the AV-ASR domain specifically, multiple works use pre-extracted visual features to improve audio-only ASR~\cite{Miao2016OpenDomainAS}. Early work~\cite{Miao2016OpenDomainAS}
leverages objects and places features from visual classifiers by 
projecting them to the same space as the audio features in a process known as Visual Adaptive Training (VAT).~\cite{Gupta2017VisualFeatures} also uses similar features, but adopts them as the beginning token of each sentence in a language modelling framework.~\cite{Caglayan2019VAT} also uses VAT, but for a sequence to sequence model. Unlike these works which use a single visual feature, we show that having multiple visual features improves performance. The closest to our work is LLD~\cite{Ghorbani2021LLD}, which also uses a stream of visual features extracted from the MIL-NCE model~\cite{miech2020end}. Their fusion method, however, consists of a complicated deliberation decoder, and while they initialize their model with audio-only pretraining, they then finetune the entire audiovisual model end-to-end. In contrast, most of our model remains frozen, and only lightweight adapters are tuned on a small amount of audio-visual data.    
All previous works are also only focused on the instructional video domain, reporting results either on internally collected datasets or the publicly released How2~\cite{sanabria2018how2}. Our focus instead is on zero-shot generalisation across multiple domains, including audio-only Librispeech~\cite{panayotov2015librispeech} (from audiobooks) and Ego4D~\cite{grauman2022ego4d} (egocentric video). We believe this is a more useful setting for actual deployment of such models.

\begin{figure*}
    \centering
    \includegraphics[width=1\linewidth]{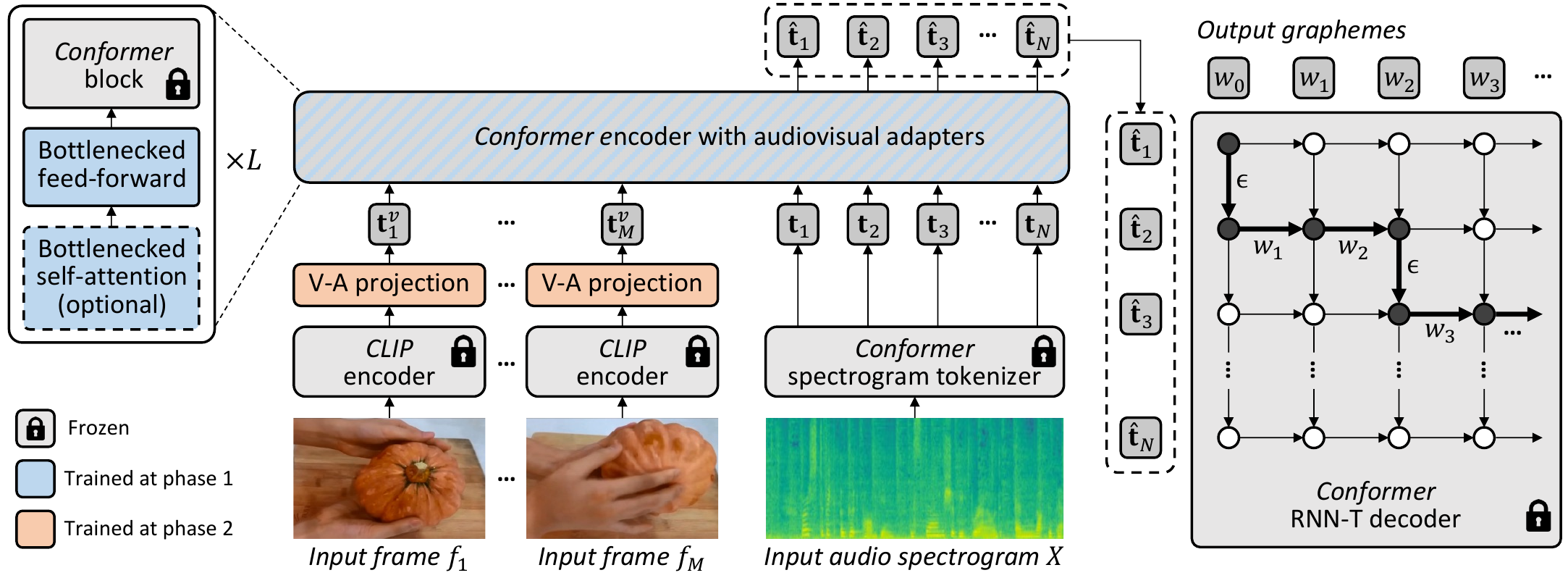}
    \caption{\textbf{Overall architecture and training procedure for \model.} Our architecture consists of a frozen Conformer encoder-decoder model~\cite{chiu2022self}, and a frozen CLIP~\cite{Radford2021CLIP} encoder (frozen layers shown in grey with a lock symbol), in conjunction with two lightweight trainable modules - (i) visual projection layer (orange) and bottleneck adapters (blue) to enable multi-modal domain adaptation. We propose a two-phase curriculum learning strategy - the adapters (blue) are first trained without any visual tokens, after which the visual projection layer (orange) are tuned while all the other parts are kept frozen.}
    \label{fig:architecture}
\end{figure*}

\section{Method}
\label{sec:method}
Unlike previous AV-ASR works which test only on instructional videos~\cite{Miao2016OpenDomainAS,Gupta2017VisualFeatures,Palaskar2018EndtoendMS,Caglayan2019VAT,Srinivasan2020LookingEL}, our goal is \textit{zero-shot} generalization across multiple AV domains, while still maintaining good performance on traditional audio-only benchmarks. To do this, we start with an \textit{existing} state-of-the-art ASR model, and adapt it for unconstrained AV-ASR. Visual features are obtained from a strong pretrained visual model, and added to the model via the following two components - (i) we linearly project visual features into the audio token embedding space, and (ii) we inject lightweight adapters into the encoder of the frozen ASR model to allow domain adaptation.
During training, we only tune these two sets of additional parameters, while both the ASR model and the visual feature extractor are~\textit{frozen} (see Figure~\ref{fig:architecture}). 

We do this because there are two forms of adaption that are required here - (i) adapting to new video domains and (ii) adapting to multimodal input, both of which we would like to do \textit{without} catastrophic forgetting. 
Because of the challenges with this setup, we also introduce a curriculum learning strategy to stabilize the learning process, without which the model fails to utilize the visual features effectively.
In this section, we first describe the main components of our network architecture (Sec.~\ref{sec:model}) and then introduce our zero-shot curriculum learning strategy and training loss functions (Sec.~\ref{sec:curriculum}).



\subsection{Model Architecture} \label{sec:model}
In this section we describe the key components of our architecture - (i) the frozen conformer encoder and decoder, (ii) the visual encoder and projection layers for visual feature extraction and projection, and (iii) additional adaptation layers in the backbone for audio-only domain adaptation. A diagram is show in Figure~\ref{fig:architecture}. 
\subsubsection{Frozen Conformer ASR Model}
We start with a frozen ASR model that achieves state-of-the-art performance on traditional ASR benchmarks~\cite{panayotov2015librispeech}. 
Specifically, we use BEST-RQ~\cite{chiu2022self} that adopts a Conformer~\cite{gulati2020conformer} model with an RNN-Transducer (RNN-T)~\cite{graves2012sequence}.
The model is pretrained on LibriLight~\cite{kahn2020libri} in a self-supervised manner using a random projection quantization technique, after which it is then finetuned for ASR on LibriSpeech~\cite{panayotov2015librispeech} using supervised training. The conformer consists of convolution-augmented transformer blocks (conformer blocks), which operate on audio token features that are extracted from a spectrogram via a stack of convolution and linear layers~\cite{gulati2020conformer}.
BEST-RQ uses ConformerXL as a backbone, which has 0.6B parameters~\cite{zhang2020pushing} -- note that training such a large model end-to-end is extremely compute heavy -- and requires a large pretraining dataset (made possible by self-supervised learning on LibriLight). This self-supervised training also enables the model to generalize well across numerous domains. 
After pretraining, an RNN-T decoder is added to  Conformer to generate text output for ASR with 1,024 WordPiece tokens~\cite{wu2016google}.
The RNN-T decoder generates a sequence of tokens consisting of grapheme tokens or a special output token, which represents moving to the next input token (See Figure~\ref{fig:architecture}, right for a diagram of the decoder).

Formally speaking, given the log-mel spectrogram $\mathbf{X}\in\mathbb{R}^{\hat{N}\times S}$ with $S$ mel spectrogram bins in a length of $\hat{N}$ converted from the input audio waveform, the tokenizer outputs a set of audio tokens $\{\mathbf{t}_i\}_1^N = h_\mathrm{tok}(\mathbf{X})$ where $D$ is the token embedding dimensionality and $N=\hat{N}/4$.
The encoder then contextualizes the audio tokens through a series of conformer blocks, each of which is a stack of feed-forward, multi-head self-attention, convolution layers followed by another feed-forward layer. 
The output of each layer is added with a residual connection.
This process produces $N$ contextualized tokens $\hat{\mathbf t}_i\in\mathbb{R}^{D}$, \ie, $\{\hat{\mathbf t}_i\}_1^N = h_\mathrm{enc}(\{\mathbf{t}_i\}_1^N)$.
The decoder finally generates the transcripts by predicting a sequence of $K$ graphemes from the contextualized audio tokens.
Given a token $\hat{\mathbf t}_i$ and previously generated grapheme $w_{j-1}$, the decoder generates the next grapheme $w_{j} = h_{\mathrm{dec}}(\hat{\mathbf t}_i, w_{j-1})$ where $w_j \in \mathcal{V} \cup \{\epsilon\}$ with the vocabulary of the predefined graphemes $\mathcal{V}$ and a special blank token $\epsilon$ that represents moving to the next token $\hat{\mathbf t}_{i+1}$ in the generation process.
The decoder $h_\mathrm{dec}$ is implemented as a two layer LSTM module with a grapheme classification head.
Note that at a single audio token index $i$, multiple graphemes can be emitted (vertical arrows) until an $\epsilon$ is emitted (horizontal arrows) as depicted in Figure~\ref{fig:architecture}.

\subsubsection{Lightweight Adapters}
In order to enable domain adaption in the model, we interleave an adapter layer within each conformer block of the encoder. 
Note that the BEST-RQ model has strong generalization capability, which we want to maintain. 
Hence we design our adapters to be lightweight, to prevent drastic domain shift and catastrophic forgetting.
Given $N$ audio tokens ${\mathbf t}_i$ and $M$ projected visual tokens $\mathbf{t}_j^{v}$ (which will be described next) at a certain layer $l$\footnote{The layer index $l$ is omitted for notational simplicity.}, we compute the adapted token features $\tilde{\mathbf t}_i$ and $\tilde{\mathbf{t}}_j^v$ using an adapter layer by $\{\tilde{\mathbf t}_i\} \cup \{\tilde{\mathbf{t}}_j^v\}=\mathrm{adapt}(\{{\mathbf t}_i\} \cup \{{\mathbf{t}}_j^v\};\phi)$ where $\mathrm{adapt}(\cdot)$ is an adapter layer parameterized by $\phi$.
We introduce and experiment with the following two types of lightweight adapters: \\
\noindent\textbf{Feed-forward Adapters (\textbf{FF}).}
The simplest design is to independently project each token.
To achieve this, we use a two-layered MLP with a residual connection as our adapter.
To make the layer lightweight, we set the dimensionality of the hidden layer to $B$, where $B \ll D$. This allows the adaptor to effectively act as a bottleneck, and reduces total additional parameters. \\
\noindent\textbf{Feed-foward Adapters with Self-Attention (FF+SA).}
The feed-forward adapters described above operate independently for each token. We can perform an additional contextualization across the input tokens via a self-attention layer~\cite{vaswani2017attention}.
To reduce additional parameters, we apply the same bottleneck projection technique as before, where each input token is transformed into a $B$ dimensional query, key and value for attention, after which the attended feature is projected back into the $D$ dimensional feature space. 
For multi-head self-attention, each head projects features into $B/H$ dimensional spaces instead where $H$ stands for the number of heads.
This module is used with a residual connection and a feed-forward module described above; the combination of these forms a transformer block with bottlenecks.
While this (FF+SA) allows additional contextualization across tokens, it introduces four times more parameters than vanilla FF adapters.

\subsubsection{Visual Feature Extraction and Projection}
Given a sequence of $M$ video frames $\mathbf f_i$, we extract a $\hat{D}$ dimensional visual feature $\mathbf v_i = g(\mathbf f_i)$ per frame using a pretrained visual encoder $g$.
Specifically, we use the CLIP encoder~\cite{radford2021learning} with ViT-L/14~\cite{dosovitskiy2020image} as our visual backbone, which is known to have strong zero-shot generalization capability~\cite{radford2021learning}.
Because the CLIP encoder is frozen, we add a linear layer\footnote{We tested more complex MLP projectors and found that a single linear layer is sufficient for good performance as detailed in the appendix.} to project the visual features into the audio token embedding space, \ie, $\mathbf t^v_i = \mathrm{proj}(\mathbf v_i;{\theta})$ where $\mathbf t^v_i \in \mathbb{R}^{D}$ and $\theta$ is a set of the parameters in the projection layer.
The projected visual tokens are fed to the Conformer encoder together with audio tokens ${\mathbf t}_i$. Note that these visual projection layers are essentially performing a type of prompt tuning~\cite{lester-etal-2021-power,mokady2021clipcap} since the rest of the ASR model is frozen.

\subsection{Training Strategy} \label{sec:curriculum}
It is a well-known that AV-ASR is an audio-dominant task, which is why previous works are forced to devise training strategies that prevent the audio stream from dominating training~\cite{gabeur2022avatar}.
We observe a similar phenomenon while jointly training both sets of additional parameters (adapters and visual projections). The visual information is not used (similar performance with and without), and training is dominated by the model only adapting to the finetuning \textit{audio} domain. We hence introduce a curriculum training strategy. We first describe our finetuning data, the loss function, and then the curriculum in the next few paragraphs. \\
\noindent\textbf{Zero-shot Training with Web Videos.}
Our extended model has two sets of new parameters $\theta$ and $\phi$ introduced for the visual projection layer and the adapters respectively.
Since it is labor-intensive and costly to collect new training benchmarks for AV-ASR, we train these new parameters without manually labeled data.
We use unlabeled web videos online along with the outputs of an ASR model as pseudo ground truth.
Our goal is to aid the pretrained ASR model with visual understanding capability using only these automatically collected transcripts; the trained model is then tested in a zero-shot setting on manually annotated public AV-ASR benchmarks.

\noindent\textbf{Loss Function.}
As the RNN-T decoder in the pretrained ASR model is kept frozen in \model, we adopt the same loss function that is used for ASR pretraining.
With an RNN-T decoder, the probability of a transcript $W=\{w_1, w_2,\cdots, w_K\}$ is obtained by marginalizing the probabilities of all valid generation paths $y$ (\eg, the path with bold arrows in Figure~\ref{fig:architecture}), \ie,
\begin{align}
    P(W|X)=\sum_{y\in\mathcal{Y}}{\prod_{(i,j)\in y}}P(w_j|\hat{\mathbf{t}}_i,w_{0:j-1})
\end{align}
where $\mathcal{Y}$ is a set of all valid paths $y$ (paths on the grid from $(0,0)$ to $(N+1,K)$ in Figure~\ref{fig:architecture}) which is a sequence of pairs of token and output grapheme indices $(i, j)$, and $P(w_j|\hat{\mathbf{t}}_i,w_{0:j-1})$ is estimated by our decoder $h_\mathrm{dec}(\hat{\mathbf{t}}_i,w_{j-1})$.
We train our model by minimizing the negative log-likelihood of the pseudo-GT transcripts $\hat{W}$ of input videos:
\begin{align}
    \mathcal{L}(\theta, \phi)=-\sum_i\log{P(\hat{W}_i|X_i; \theta,\phi)}.
\end{align}

\noindent\textbf{Curriculum Learning for Visual Processing.}
We discover empirically that with a naive first round of joint training, our model struggles to learn both the adapters and the visual projectors in one go (as shown in the experiments, the issue becomes more severe as more visual tokens are added).
To mitigate this issue, we propose a two-phase curriculum learning strategy that decouples these two factors (domain adaption and visual feature integration) and trains the network in a sequential manner.
In the first phase, the adapter parameters $\phi$ are optimized using $\argmax_{\phi}\mathcal{L}(\theta,\phi)$ as an objective.
Note that at this phase, we do not feed visual tokens at all and thus $\theta$ is an empty set.
Once $\phi$ is trained, we add the visual tokens and train the visual projection layers $\theta$ using $\argmax_{\theta}\mathcal{L}(\theta,\phi)$.
During this second phase of training, $\phi$ is kept frozen.

The first stage focuses on audio domain adaptation. By the second phase, the adapters are completely frozen and the visual projector must simply learn to generate visual prompts that project the visual tokens into the audio space. 
In this way, our curriculum learning strategy allows the model to incorporate visual inputs as well as adapt to new audio domains in AV-ASR benchmarks.
We apply each phase just once, as an iterative application of alternating phases leads to performance degradation. This is further discussed in the appendix.

\noindent\textbf{Content Word Masking.}
We adopt the content word masking from \cite{gabeur2022avatar} to encourage the models to further focus on visual understanding.
We observe that the original zero-padded masking introduced in~\cite{gabeur2022avatar} causes instabilities and therefore we add Gaussian noise to the audio input corresponding to masked words, which stabilizes optimization.

\section{Experiments}
\label{sec:experiments}


\begin{figure*}
    \centering
    \includegraphics[width=\linewidth]{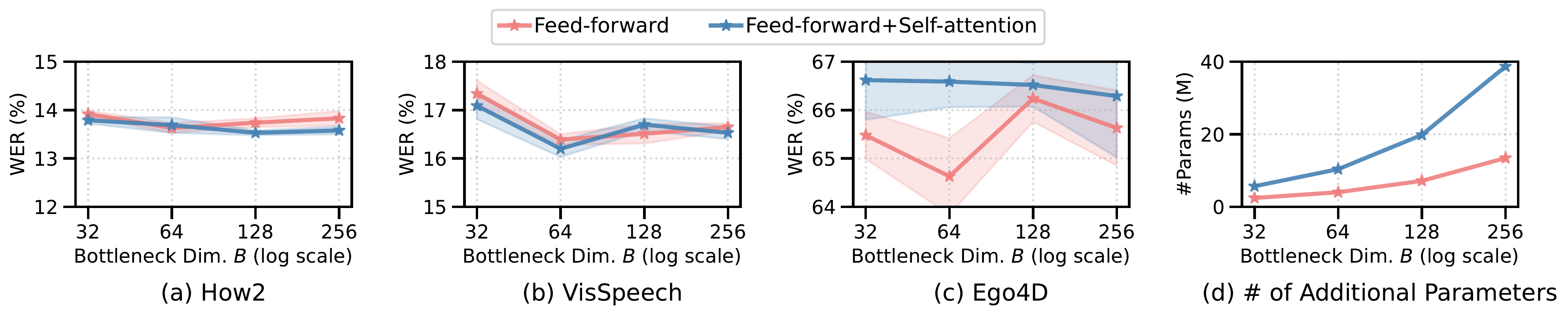}
    \vspace{-0.6cm}
    \caption{\textbf{Effects of different architectures (feed-forward (FF) vs feed-forward + self-attention (FF+SA)) and the bottleneck dimensionality $B$ of adaptor layers on performance.} Results are for audiovisual models trained with our curriculum learning, and are shown on 3 datasets in the zero-shot setting (lower WER\% is better). We show that a bottleneck dimension of 64 with FF layers achieves the best or almost the best performance (a,b,c) with the least number of additional parameters (d). Best viewed in color.}
    \label{fig:type_bottleneck}
    \vspace{-0.1cm}
\end{figure*}


\begin{figure*}
    \centering
    \includegraphics[width=\linewidth]{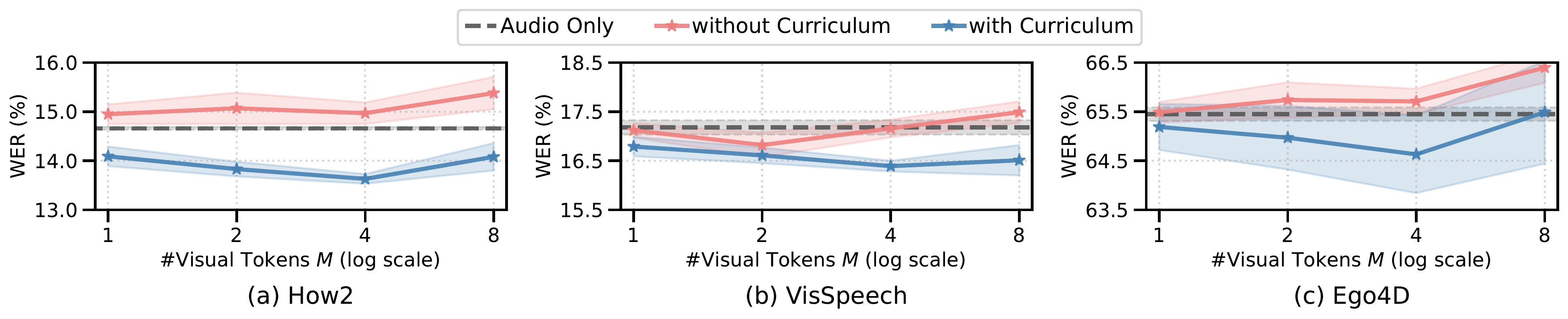}
    \vspace{-0.6cm}
    \caption{\textbf{Effects of curriculum learning and the number of visual tokens $M$ on performance.} Red and blue lines are for audiovisual models and are shown on 3 datasets in the zero-shot setting (lower WER\% is better). Using the curriculum helps on all 3 datasets (for How2 (a) and Ego4D (c) it is crucial for outperforming audio-only performance). Performance improves up until 4 visual tokens, at which point it saturates. Best viewed in color.}
    \label{fig:curriculum_vp}
    \vspace{-1em}
\end{figure*}


\subsection{Experimental Settings}
\paragraph{Implementation Details.}
As mentioned earlier, we use BEST-RQ~\cite{chiu2022self} as the frozen ASR model.
Since it has 24 conformer blocks, we add 24 adapters (one in each layer) in all experiments.
When added, all adapters and visual projectors are randomly initialized.
The decoder predicts WordPiece tokenized graphemes with a vocabulary size of 1,024.
In the adapters, we apply layer norm~\cite{ba2016layer} at every residual connection.
For both phases of training, we use standard SGD with momentum with a moving average coefficient of 0.9 and a cosine learning rate schedule; the initial learning rate is set to 0.4.
We train for 40K and 30K iterations in phase 1 and 2 respectively, with a batch size of 256 on 32 TPU v4 chips.
We run 5 independent experiments and report the mean scores for ablation studies. 
When testing Audiovisual models on audio-only benchmarks, we feed dummy visual inputs (zero tensors).

\noindent\textbf{Metrics.}
We use word error rate (WER) for all evaluation (lower is better).
The alignment between predicted words and ground truth is computed using dynamic programming.
The WER is then computed by the number of errors (deletions, substitutions and insertions) across the whole test set divided by the number of ground truth words. 

\noindent\textbf{Baselines.}
We compare \model~to two strong baselines proposed this year - (i) the state-of-the art AV-ASR model AVATAR~\cite{gabeur2022avatar} and (ii) the state-of-the-art ASR (audio only) model BEST-RQ~\cite{chiu2022self}. We apply both models to the same settings as \model~ for a fair comparison.

\subsection{Datasets}
The additional parameters in our model are finetuned on the HowTo100M dataset, which contains instructional videos from YouTube.
In order to assess generalization, we evaluate across different domains -- LibriSpeech (audiobooks), How2 and VisSpeech (YouTube instructional videos) and Ego4D (egocentric video of daily-life activities). Note that VisSpeech consists of more unconstrained video (background noise, challenging accents etc) than How2. More details for each dataset are provided below. \\
\noindent\textbf{LibriLight~\cite{kahn2020libri} and LibriSpeech~\cite{panayotov2015librispeech}.}
LibriLight is an unlabelled speech dataset that is used to pretrain BEST-RQ.
The model is then finetuned for ASR on LibriSpeech containing 960 hours audio with manually annotated GT transcripts.
For a fair comparison, we also use LibriSpeech for pretraining some of our baselines in the ablations. \\
\noindent\textbf{HowTo100M~\cite{miech19howto100m}.}
This dataset contains 1.2M instructional videos without manual annotations.
ASR is used to obtain pseudo-GT transcripts  
for training our adapters and visual projector. We remove videos present in VisSpeech and How2 (described next). \\
\noindent\textbf{How2~\cite{sanabria2018how2}.}
We use the 300hr version of How2, which consists of instructional videos with automatically collected user uploaded captions.
The videos are segmented into 5.8s short clips with 20-word long transcripts in average.
We use the validation (2,022 clips) and test (2,305 clips) splits to evaluate our model in a zero-shot setting. \\
\noindent\textbf{VisSpeech~\cite{gabeur2022avatar}.}
VisSpeech is an AV-ASR test benchmark that consists of 503 video clips with manually annotated transcripts, which are sampled from HowTo100M. The dataset curation process focuses on samples where an audio-only ASR model fails and where strong visual correlations are observed. \\
\noindent\textbf{Ego4D~\cite{grauman2022ego4d}.}
Ego4D consists of egocentric video from 74 worldwide locations and 9 countries, with over 3,670 hours of daily-life activity video. We use the audiovisual diarization benchmark in the Ego4D challenge\footnote{\url{https://ego4d-data.org/docs/challenge/}}. 
It consists of 585 5-minutes long egocentric video clips splited into train (397 clips), validation (51 clips) and test (137 clips) sets.
We report zero-shot results on the validation set as the test annotations are not released.
We evaluate transcripts on segmented clips based on GT boundaries.

\subsection{Results}
In this section, we show ablations of the various design choices in our model (adapter architecture and bottleneck dimension), and then discuss the impact of curriculum learning and the benefit of adding visual tokens (including the impact of the number of visual tokens). We then show an ablation discussing the impact of adding both adapters and visual tokens, and the impact of finetuning dataset size. Finally, we show zero-shot performance of our model compared to state of the art baselines. Note that all ablations and results are provided on all $3$ downstream datasets in a zero-shot setting -- How2, VisSpeech and Ego4D.

\begin{figure*}
    \centering
    \includegraphics[width=\linewidth]{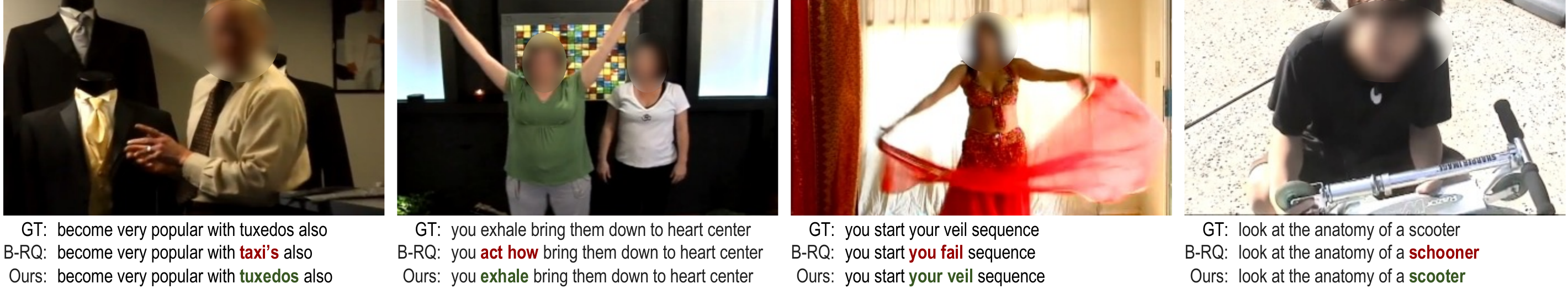}
    \includegraphics[width=\linewidth]{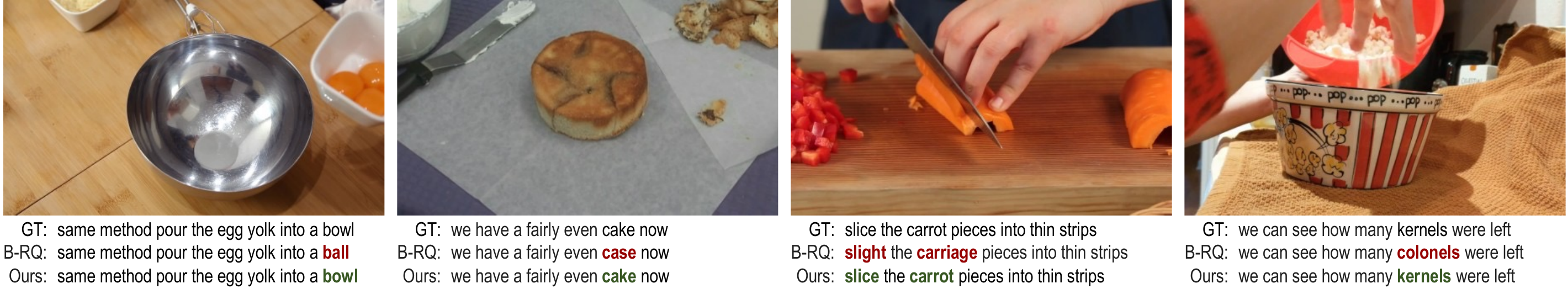}
    \includegraphics[width=\linewidth]{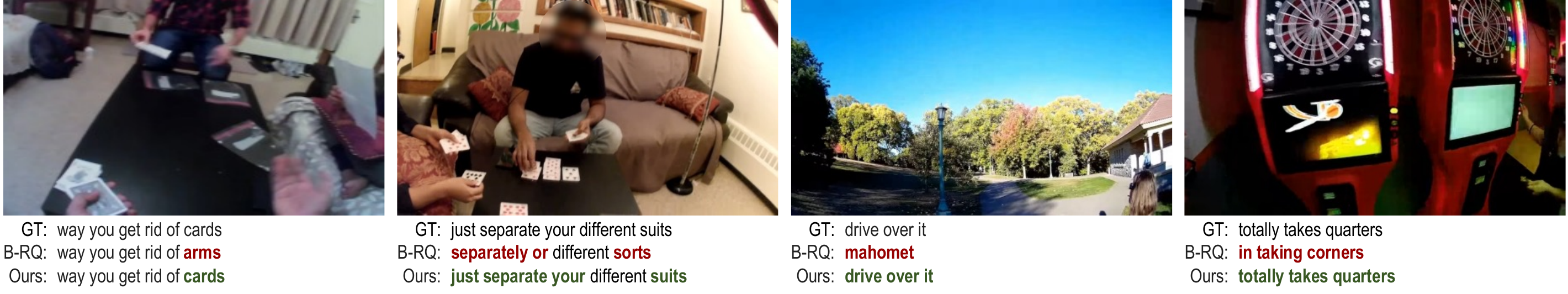}
    \vspace{-0.7cm}
    \caption{\textbf{Qualitative Results on  How2 (top), VisSpeech (middle) and Ego4D (bottom).} We show the ground truth (GT), and predictions from the audio only BEST-RQ model (B-RQ) and our audiovisual \model~(Ours) in the zero-shot setting. For each clip we show a single visual frame. Note how the visual context helps with visual objects (tuxedos, veil, scooter, bowl, cake, carrot \textit{etc}), as well as actions (exhale, drive over) and works well even in the ego-centric domain (learns driving from input of road in row 3, column 3). Errors in the predicted words compared to the GT are highlighted in red. Faces are blurred for privacy.}
    \label{fig:qualitative_results}
    \vspace{-1em}
\end{figure*}

\noindent\textbf{Adapter Architecture and Bottleneck Dimensionality.}
Figure~\ref{fig:type_bottleneck} compares results with feed-forward adapters (FF) only vs adapters with both feed-forward and self-attention (FF+SA). We also vary the bottleneck dimension from 32 to 256.
We observe that on How2 (Figure~\ref{fig:type_bottleneck}a) and VisSpeech (Figure~\ref{fig:type_bottleneck}b), both adapter types perform similarly although FF+SA uses significantly more parameters than FF (Figure~\ref{fig:type_bottleneck}d), indicating that a simple projection is enough for strong adaptation. 
On Ego4D (Figure~\ref{fig:type_bottleneck}c), simple FF outperforms FF+SA by a large margin, potentially because of the larger domain gap (instructional edited videos online to egocentric daily activity videos). The greater number of parameters in FF+SA may result in a larger shift to the instructional video domain and away from Ego4D. 
Figure~\ref{fig:type_bottleneck} also shows the effect of different bottleneck dimensions. In general the WER comes down from $32$ to $64$, but saturates at $B=64$ across all datasets with FF, while introducing only few additional parameters (0.6\% of the number of parameters in BEST-RQ).
Hence in the rest of experiments, we adopt FF adapters with $B=64$.
\begin{table}[t]
    \centering
    \caption{\textbf{Effect of visual tokens (VT) and adapter layers.}
    Results on 3 datasets are obtained in the zero-shot setting (lower WER\% is better). 
    The first row corresponds to the vanilla pretrained BEST-RQ. 
    Visual projector is added only when feeding VT. 
    The gains from both VT and adapters are complementary.}
    \label{tab:model_ablations}
    \vspace{-0.1cm}
    \scalebox{0.85}{
    \begin{tabular}{cc|ccc}
        \toprule
        \bf VT & \bf Adapters & \bf How2 & \bf VisSpeech & \bf Ego4D \\ \midrule
         & & 21.90 & 31.61 & 77.98 \\
        $\checkmark$ & & 19.74 \small{$\pm$ 0.04} & 31.13 \small{$\pm$ 0.06} & 76.50 \small{$\pm$ 0.11}\\
        & $\checkmark$ &  14.66 \small{$\pm$ 0.03} & 17.18 \small{$\pm$ 0.15} & 65.45 \small{$\pm$ 0.14} \\
        $\checkmark$ & $\checkmark$ & 13.63 \small{$\pm$ 0.10} & 16.39 \small{$\pm$ 0.11} & 64.63 \small{$\pm$ 0.79} \\\bottomrule
    \end{tabular}
    }
    \vspace{-1em}
\end{table}
\begin{table}[t]
    \centering
    \caption{\textbf{Effect of training dataset size.} Results are for audiovisual models trained with our curriculum learning, and are shown on 3 datasets in the zero-shot setting (lower WER\% is better).
    Only 5\% of HowTo100M is required.}
    \label{tab:dataset_size}
    \vspace{-0.1cm}
    \scalebox{0.85}{
    \begin{tabular}{c|ccc}
        \toprule
        \bf Training-set size & \bf How2 & \bf VisSpeech & \bf Ego4D \\\midrule
        5\% & 13.69 \small{$\pm$ 0.17} & 16.60 \small{$\pm$ 0.17} & 64.75 \small{$\pm$ 1.05} \\
        100\% & 13.63 \small{$\pm$ 0.10} & 16.39 \small{$\pm$ 0.11} & 64.63 \small{$\pm$ 0.79} \\
        \bottomrule
    \end{tabular}
    }
    \vspace{-1em}
\end{table}
\begin{table*}[t]
    \centering
    \caption{\textbf{Comparison to state-of-the-art methods for zero-shot performance across different AV-ASR datasets.} We also show performance on LibriSpeech which is audio-only. Results are reported as WER \% (lower is better). Note that AVATAR and BEST-RQ are finetuned end-to-end (all parameters) on HowTo100M, whereas for our model, only the visual projectors (VP) and adapters are finetuned on 5\% of the dataset. PT means pretraining. When a model is marked with both LibriSpeech and HowTo100M pretraining, we first train the model on LibriSpeech and then on HowTo100M next. For LibriSpeech evaluation, we report numbers on test-clean set. *LibriSpeech trained model is evaluated directly on LibriSpeech test set. }
    \vspace{-0.2cm}
    \label{tab:zeroshot}
    \scalebox{0.85}{
    \begin{tabular}{c|c|ccc|c|ccc}
        \toprule
         &   &   & \multicolumn{2}{c|}{\textbf{HowTo100M PT}} & &  &  &  \\ 
       \textbf{Method} & \textbf{Modality} & \textbf{LibriSpeech PT} & Pretrained params & Data \% & \textbf{LibriSpeech} & \textbf{How2} & \textbf{VisSpeech} & \textbf{Ego4D} \\ \midrule
        AVATAR~\cite{gabeur2022avatar} & A & $\checkmark$  & -- & -- & 8.85 & 39.43 & 65.33 & 110.86 \\
        AVATAR~\cite{gabeur2022avatar} & A+V & -- & All & 100 & 24.65 & 17.23 & 35.66 & 92.03 \\
        AVATAR~\cite{gabeur2022avatar} & A+V & $\checkmark$ & All & 100 & 24.08 & 18.37 & 35.59 & 71.97 \\ \midrule 
        BEST-RQ~\cite{chiu2022self} & A & $\checkmark$ & -- & -- & 1.60* & 21.90 & 28.62 & 77.98 \\
        BEST-RQ~\cite{chiu2022self} & A & $\checkmark$ & All & 100 & 5.60 & 15.32 & 16.69 & 68.34 \\ \midrule
        \bf \model~(Ours) & \bf A+V & \bf $\checkmark$ &  VP + Adapters & 5 & 4.36 & \bf 13.69 & \bf 16.60 & \bf 64.75 \\ \bottomrule
    \end{tabular}
    }
    \vspace{-0.2cm}
\end{table*}

\noindent\textbf{Curriculum Learning and Visual Tokens.}
We show the results of \model~with and without the proposed two-stage curriculum in Figure~\ref{fig:curriculum_vp}, and also compare to an audio-only baseline which had only FF adapters with $B=64$ and no visual information.
Without curriculum learning, our AV-ASR model is worse than the audio-only baseline across all datasets, with the gap increasing as more visual tokens are added. 
In contrast, when the proposed two-phase curriculum is applied, our AV-ASR model performs significantly better than the baseline audio-only model.
We also test our model with different number of visual input tokens (where one token corresponds to one frame).
More visual tokens improves the model up until $M=4$ with up to 7.0\% relative improvement, after which performance begins to degrade. Hence we set $M=4$ in all experiments. \\
\noindent\textbf{Complementary Gain of Additional Components.}
Table~\ref{tab:model_ablations} shows the effect of our additional lightweight components (projection layer for visual tokens and adapter layers) for zero-shot AV-ASR. The first row is simply the vanilla baseline (frozen BEST-RQ).
We observe that adding projected visual tokens and adapters bring individual gains to the baseline (the former adding visual information and the latter aiding with audio-domain adaptation), and when  combined with our curriculum learning, are complementary to performance, achieving the lowest WER. \\
\noindent\textbf{Training Dataset Size}.
Given our additional components are so lightweight, we test whether adaptation can be done with a small amount of weakly labelled data. 
The results in Table~\ref{tab:dataset_size} show only 5\% of HowTo100M training data performs on par with the full dataset -- \ie\ the pretrained knowledge in BEST-RQ and CLIP yields considerable data efficiency to the model. Ablation results with more data fractions are provided in the appendix. \\
\noindent\textbf{Comparisons to Zero-shot Baselines on AV-ASR.}
We compare our model to baselines in Table~\ref{tab:zeroshot}, for zero-shot performance on all 3 AV-ASR benchmarks\footnote{Note that the original AVATAR and BEST-RQ papers do not report this. We apply these models in the same setting as ours for a fair comparison.}
\model~outperforms AVATAR and BEST-RQ on all, even outperforming both AVATAR and BEST-RQ when they are fully finetuned on LibriSpeech and then 100\% of HowTo100M (3rd and 5th row). Note for BEST-RQ, this involves finetuning 0.6B params. Our model, in contrast only finetunes 4M params on 5\% of HowTo100M. 

\noindent\textbf{Comparisons to Zero-shot Baselines on LibriSpeech.}
Even though this is not the main goal of this work, we also investigate performance on LibriSpeech, which is audio-only (Table~\ref{tab:zeroshot}). Note other AV-ASR works do not do this, but we believe it is important for deployment of AV-ASR models. We first note that AVATAR pretrained on LibriSpeech and then finetuned on HowTo100M performs poorly when re-evaluated on LibriSpeech (showing severe catastrophic forgetting between rows 1 and 3). We believe this is because all parameters are trained end-to-end. On the other hand, AVFormer performs much better on LibriSpeech (4.36 vs 24.08), and is much closer to BEST-RQ's 1.60 which is a model tuned only for LibriSpeech and incapable of AV-ASR, while \model~achieves SOTA on AV-ASR as well. \\
\noindent\textbf{Qualitative Results.} Qualitative examples are provided in Fig. \ref{fig:qualitative_results} comparing our method to audio-only BEST-RQ for zero-shot ASR. We show that for all 3 downstream AV-ASR datasets, visual context improves mistakes that are made on objects (\textit{eg.} tuxedos, veil and scooter from the top row), actions (exhale - top row, second column), and even corrects a homophone \footnote{same pronunciation, different spelling} (colonels to kernals, row 2 column 4). \\
\begin{table}[t]
    \centering
    \caption{\textbf{Finetuning performance on How2 and Ego4D.} We outperform all previous works on How2 that use frozen visual features. AVATAR is trained end-to-end, with all visual parameters finetuned. Scores are in WER~\%.
    }
    \label{tab:sota_comp}
    \scalebox{0.85}{
    \begin{tabular}{lc|cc}
        \toprule
        \bf Method & \bf Frozen visual\ feats & \bf How2 
        & \bf Ego4D \\\midrule
        VAT~\cite{Caglayan2019VAT} &$\checkmark$ &  18.0 & -- \\
        MultiRes~\cite{paraskevopoulos2020multires} & $\checkmark$ & 20.5 & --\\
        LLD~\cite{Ghorbani2021LLD} &$\checkmark$ &  16.7 & -- \\
        AVATAR \cite{gabeur2022avatar} & & 9.11  & 55.27 \\ \midrule
        \model~(Ours) & $\checkmark$ & 10.22 & \textbf{55.23} \\
        \bottomrule
     \vspace{-2em}
    \end{tabular}
    }
\end{table}
\noindent\textbf{Comparisons to SOTA after Finetuning.}
For completeness, we also show finetuning results on two domains - instructional (How2) and egocentric (Ego4D) videos in Table \ref{tab:sota_comp}. We outperform all previous works on How2 that use frozen visual features. Our model is also not too much worse (How2) or on par (Ego4D) with AVATAR, even though AVATAR is trained end-to-end, and all parameters (including a large visual encoder) are finetuned. 

\section{Conclusion}
\label{sec:conclusion}
We present AVFormer, a lightweight method for adapting existing, frozen state-of-the-art ASR models for AV-ASR. Our approach is practical and achieves impressive zero-shot performance. 
As ASR models get larger and larger, tuning the entire parameter set of pre-trained models becomes impractical for different domains. Our method seamlessly allows both domain transfer and visual input mixing in the same, parameter efficient model. 


{\small
\bibliographystyle{ieee_fullname}
\bibliography{egbib}
}

\clearpage

\appendix

\part{Appendix} 
In this appendix we provide additional ablations with varying the number of adapter layers and with more complex visual projectors in Section~\ref{sec:num_layers} and \ref{sec:complex_vp}, respectively.
In Section~\ref{sec:iter_train}, we investigate the effects of iterative training.
Then we supplement Table 2 of the main paper by providing results with more training dataset fractions in Section~\ref{sec:data_size},
and show additional experiments replacing the RNN-T decoder with a cross-attentional transformer decoder in Section~\ref{sec:decoder}.
Finally, we present a failure case in Section~\ref{sec:failure}.

\section{Number of Adapter Layers}
\label{sec:num_layers}

\begin{figure}[ht]
    \centering
    \begin{subfigure}[b]{\linewidth}
        \centering
        \includegraphics[width=0.75\linewidth]{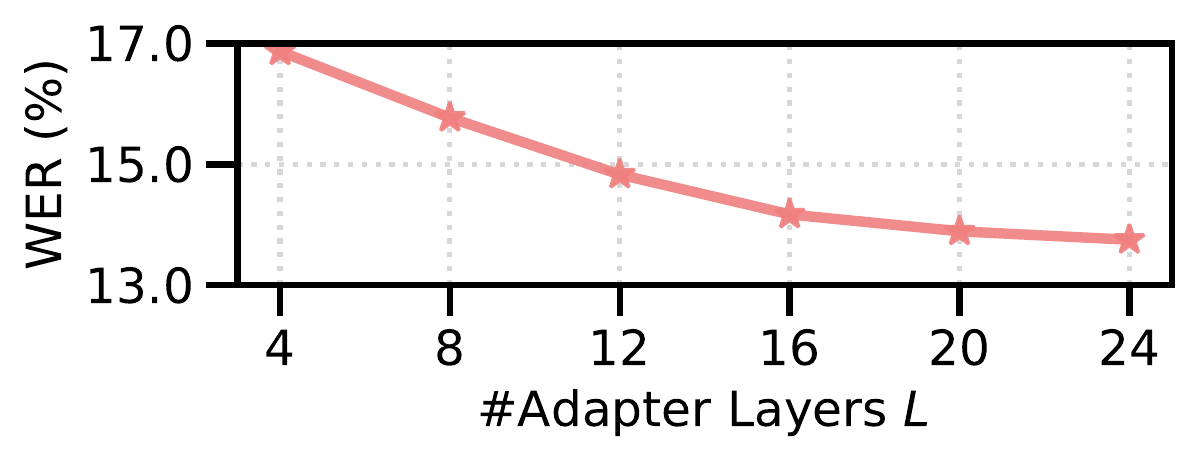}
        \caption{How2}
        \label{fig:num_l_how2}
    \end{subfigure}
    \begin{subfigure}[b]{\linewidth}
        \centering
        \includegraphics[width=0.75\linewidth]{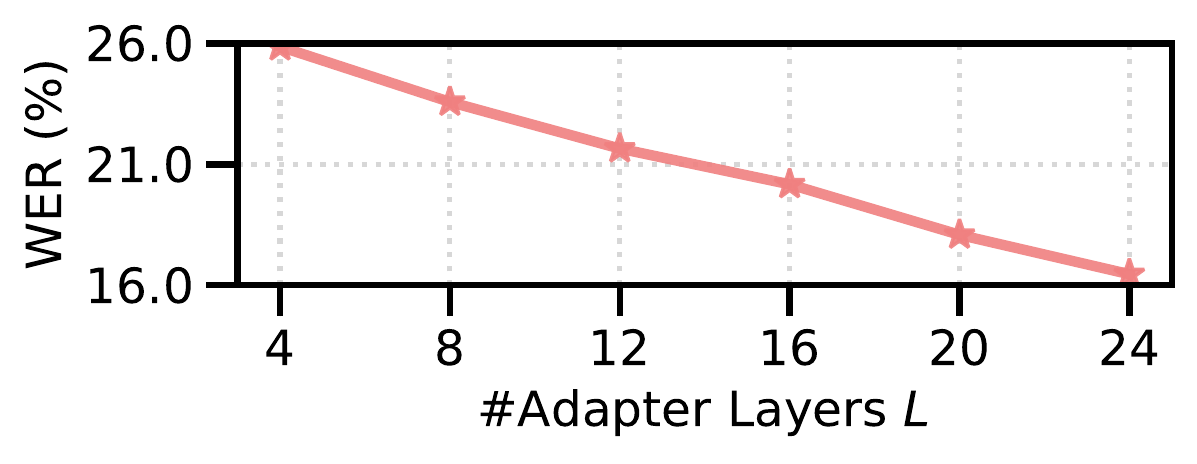}
        \caption{VisSpeech}
        \label{fig:num_l_visspeech}
    \end{subfigure}
    \begin{subfigure}[b]{\linewidth}
        \centering
        \includegraphics[width=0.75\linewidth]{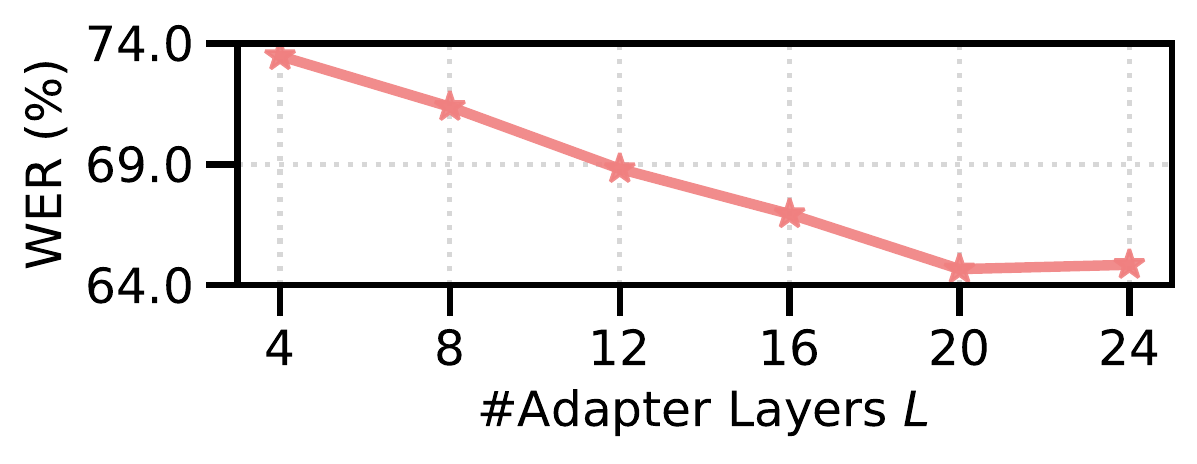}
        \caption{Ego4D}
        \label{fig:num_l_ego4d}
    \end{subfigure}
    \caption{\textbf{Effect of the number of adapter layers.} Models are trained with 4 visual tokens using our curriculum learning strategy. Performance improve on all datasets as we increase the number of adapter layers. Lower WER is better.}
    \label{fig:num_layers}
\end{figure}

Figure~\ref{fig:num_layers} shows the word error rate (WER) when varying the number of adapter layers from 4 to 24.
Note that the number of the conformer blocks in BEST-RQ is 24 and therefore, the model with 24 adapters means that an adapter is added to every conformer block in the model.
We also note that we add adapter layers to the last conformer blocks (before the decoder) when fewer than 24 layers are added to achieve the best performance.

WER for How2 (Figure~\ref{fig:num_l_how2}) and VisSpeech (Figure~\ref{fig:num_l_visspeech}) monotonically decreases as we add more adapter layers to the model.
For Ego4D (Figure~\ref{fig:num_l_ego4d}), the performance saturates at 20 layers.
These results suggest that it is critical to inject an adapter into every conformer block.

\begin{figure*}[t]
    \centering
    \includegraphics[width=\linewidth]{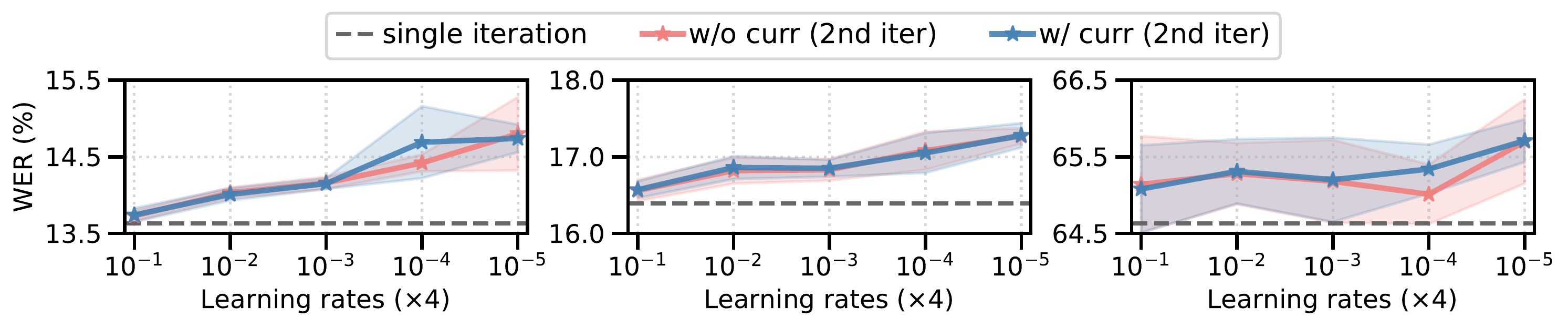}
    \caption{\textbf{Effects of iterative training on How2 and Ego4D.} Our model is trained for the second time with both or without our proposed curriculum using different learning rates.
    }
    \label{fig:iter_train}
\end{figure*}
\section{More Complex Visual Projector}
\label{sec:complex_vp}
We also test a more complex visual projector in the form of a multi-layer perceptron (MLP) with varying number of layers (Table~\ref{supptab:vp_layers}).
The results consistently show on all three datasets that a single linear layer is sufficient for good performance (lower is better), and adding more layers makes a marginal impact (within error bars). Note that similar results are observed in \cite{eichenberg2021magma} for prefix matching tasks.

\section{Iterative Training}
\label{sec:iter_train}
In this section, we investigate iterative applications of our curriculum. We train our model for the second time, both with or without our proposed curriculum. The results in Figure~\ref{fig:iter_train} present performance degradation compared to our model with single iteration (gray dotted lines) in both cases on all three benchmarks.
We observe that a larger learning rate increases the WER. We believe that this phenomenon is due to over-adaptation to HowTo100M.

\section{Effect of Dataset Size}
\label{sec:data_size}
We extend the ablation presented in Table 2 of the main paper in Table~\ref{supptab:dataset_size}.
Due to the strong pretrained knowledge in BEST-RQ, we show that only a small fraction (5\%) of the HowTo100M training dataset is enough to achieve comparable performance with training on the full dataset. This shows that our adapted model is extremely data efficient.

\begin{table}[t]
    \centering
    \scalebox{0.85}{
    \begin{tabular}{c|ccc}
        \toprule
        \bf \# layers & \bf How2 & \bf VisSpeech & \bf Ego4D \\ \midrule
         \bf 1 & \bf 13.63 \small{$\pm$ 0.10} & \bf 16.39 \small{$\pm$ 0.11} & \bf 64.63 \small{$\pm$ 0.79} \\
         2 & 13.77 \small{$\pm$ 0.09} & 16.47 \small{$\pm$ 0.34} & 64.75 \small{$\pm$ 0.81} \\
         3 & 13.93 \small{$\pm$ 0.21} & 16.49 \small{$\pm$ 0.14} & 65.20 \small{$\pm$ 0.56} \\
         4 & 13.72 \small{$\pm$ 0.11} & 16.49 \small{$\pm$ 0.25} & 65.04 \small{$\pm$ 0.50} \\\bottomrule
    \end{tabular}
    }
        \caption{\textbf{Effect of the number of MLP layers in the visual projector.} ReLU is used as the intermediate activation function.}
    \label{supptab:vp_layers}
    \vspace{-0.3cm}
\end{table}

\begin{table}[t]
    \centering
    \scalebox{0.85}{
    \begin{tabular}{c|ccc}
        \toprule
        \bf Dataset Size & \bf How2 & \bf VisSpeech & \bf Ego4D \\ \midrule
5\%&	13.69 & 16.60 & 64.75\\
10\%&	13.79 & 16.56 & 65.37\\
25\%&	13.60 & 16.57 & 64.29\\
50\%&	13.63 & 16.53 & 64.63\\
75\%&	13.66 & 16.69 & 65.11\\
100\%&	13.63 & 16.39 & 64.63\\
        \bottomrule
    \end{tabular}
    }
    \caption{\textbf{Effect of training dataset size.} Models are trained with 4 visual tokens using our curriculum strategy. Models are trained with varying fractions of HowTo100M. All scores are in WER\% (lower is better). The results show that 5\% of dataset is enough to achieve state-of-the-art performance.}
    \label{supptab:dataset_size}
\end{table}

\section{Autoregressive Decoder}
\label{sec:decoder}
Finally, we test our method with different decoders: an RNN-Transducer (RNN-T) and a transformer decoder using cross-attention (Cross-attention) introduced in \cite{vaswani2017attention}.
RNN-T is the decoder used in the pretrained BEST-RQ model, we keep the weights frozen when training for AV-ASR.
The cross-attention decoder performs autoregressive decoding while cross-attending to all input tokens.
We stack 8 decoder transformer blocks; the weights are randomly initialized and tuned during the AV-ASR training.

Table~\ref{tab:decoder} shows the results of these models on the four datasets (LibriSpeech and the three AV-ASR benchmarks). 

The cross-attention decoder performs worse than RNN-T on the three AV-ASR benchmarks, while performing \textit{significantly} worse on Librispeech.
Note that Cross-attention uses the entire set of input encodings for generating each output token whereas every output token is generated from a single input encoding with RNN-T.
However, the results show that maintaining the pretrained decoding knowledge in RNN-T is more important than introducing larger flexibility in a finetuned decoder.

\section{Failure Analysis}
\label{sec:failure}
\begin{figure}
    \centering
    \includegraphics[width=0.6\linewidth]{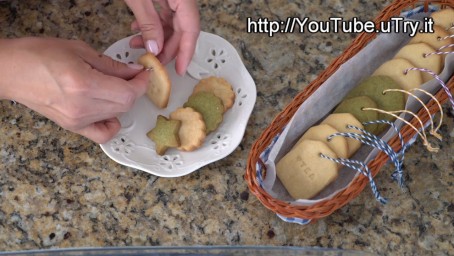}

    \vspace{0.1cm}
    \begin{tabular}{rl}
        GT: & and tie up both ends with a simple knot \\
        Ours: & and tie up both \textcolor{red}{hands}  with a simple knot
    \end{tabular}
  \caption{\textbf{A failure example on VisSpeech.}}
  \label{fig:failure}
\end{figure}

Figure~\ref{fig:failure} shows a failure case with an erroneous word `hands' introduced by the visual input. However, we find this case very rare in our extensive qualitative exploration.

\begin{table}[t]
    \centering
    \scalebox{0.85}{
    \begin{tabular}{ccccc}
        \toprule
        \bf Decoder & \bf LibriSpeech & \bf How2 & \bf VisSpeech & \bf Ego4D \\
        \midrule
        Cross-attention & 13.79 & 16.67 & 20.21 & 70.47 \\
        RNN-T & 4.40 & 13.63 & 16.39 & 64.63\\
        \bottomrule
    \end{tabular}
    }
    \caption{\textbf{Results with different decoders.} All scores are in WER\% (lower is better). RNN-T represents that an RNN-T decoder is initialized with the pretrained BEST-RQ weights and frozen. Cross-attention means that we replace the original RNN-T decoder with a autoregressive transformer decoder using cross-attention on the input token embeddings. Results are reported on all three AV-ASR benchmarks as well as on LibriSpeech.}
    \label{tab:decoder}
\end{table}

\end{document}